\documentclass{article}

\usepackage{PRIMEarxiv}

\usepackage[utf8]{inputenc} % allow utf-8 input
\usepackage[T1]{fontenc}    % use 8-bit T1 fonts
\usepackage{hyperref}       % hyperlinks
\usepackage{url}            % simple URL typesetting
\usepackage{booktabs}       % professional-quality tables
\usepackage{amsfonts}       % blackboard math symbols
\usepackage{nicefrac}       % compact symbols for 1/2, etc.
\usepackage{microtype}      % microtypography
\usepackage{lipsum}
\usepackage{graphicx}
\graphicspath{{media/}}     % organize your images and other figures under media/ folder

\usepackage{amsmath}
\usepackage{amssymb}
\usepackage{multirow}

\title{FisheyeHDK: Hyperbolic Deformable Kernel Learning for Ultra-Wide Field-of-View Image Recognition
\thanks{\textit{\underline{Preprint accepted in}}: 
\textbf{36th AAAI Conference on Artificial Intelligence, 2022.}} 
}

\author{
  Ola Ahmad \\
  Thales, Thales Digital Solutions, Canada \\
  \texttt{ola.ahmad@thalesdigital.io} \\
   \And
  Freddy Lecue \\
  Thales, Thales Digital Solutions, Canada\\
  INRIA, France\\
  \texttt{freddy.lecue@inria.fr} \\
}

\begin{document}
\maketitle

\begin{abstract}
Conventional convolution neural networks (CNNs) trained on narrow Field-of-View (FoV) images are the state-of-the art approaches for object recognition tasks. Some methods proposed the adaptation of CNNs to ultra-wide FoV images by learning deformable kernels. However, they are limited by the Euclidean geometry and their accuracy degrades under strong distortions caused by fisheye projections.
In this work, we demonstrate that learning the shape of convolution kernels in non-Euclidean spaces is better than existing deformable kernel methods.
In particular, we propose a new approach that learns deformable kernel parameters (positions) in hyperbolic space. FisheyeHDK is a hybrid CNN architecture combining hyperbolic and Euclidean convolution layers for positions and features learning. First, we provide intuition of hyperbolic space for wide FoV images. Using synthetic distortion profiles, we demonstrate the effectiveness of our approach.
We select two datasets - Cityscapes and BDD100K 2020 - of perspective images which we transform to fisheye equivalents at different scaling factors (analogue to focal lengths). Finally, we provide an experiment on data collected by a real fisheye camera. Validations and experiments show that our approach improves existing deformable kernel methods for CNN adaptation on fisheye images.
\end{abstract}

% keywords can be removed
%\keywords{First keyword \and Second keyword \and More}

\section{Introduction}
Fisheye cameras are designed with ultra-wide field of view (FoV) lenses to offer wide images. They are commonly used in many computer vision applications. In particular, autonomous vehicles heavily rely on perception tasks such as semantic segmentation and object detection from the environment surrounding the vehicle for path and motion planning, driving policy, and decision making \cite{8842620, Yogamani_2019_ICCV}. To optimize the load on autonomous vehicles, only four fisheye cameras each with FoV up to $180^\circ$ can be used to provide the total $360^\circ$ scene coverage \cite{Yogamani_2019_ICCV}. Automated drones \cite{1544998},  augmented reality \cite{7892358} and surveillance \cite{7066501} are other interesting applications of large FoV cameras. However, unlike pinhole projection models of narrow FoV cameras (where straight lines in 3D world are projected to straight lines in image plane), fisheye images suffer from non-linear distortion in which straight lines are mapped to curvilinear (distorted lines) to provide large FoV on a finite spatial support. According to \cite{kannala06}, fisheye distortion does not obey one specific projection model. A general polynomial mapping of fourth order was proven to be an accurate approximation of fisheye camera model \cite{Yogamani_2019_ICCV, yin_fisheyerecnet_2018}. This model was used to synthesize fisheye effects on perspective images. It was also used for calibration and distortion correction prior to fisheye image recognition tasks. However, estimating the parameters of the polynomial model of order four is an inverse problem which is ill-posed, in practice, under the lack of information and careful assumptions. In this paper, we address the challenge of ultra-wide FoV image recognition without the need for distortion correction. A challenge that is quite recent in computer vision research areas.
\par
\begin{figure}[t]
\centering
%\fbox{\rule{0pt}{2in} \rule{0.9\linewidth}{0pt}}
 \includegraphics[width=0.6\linewidth]{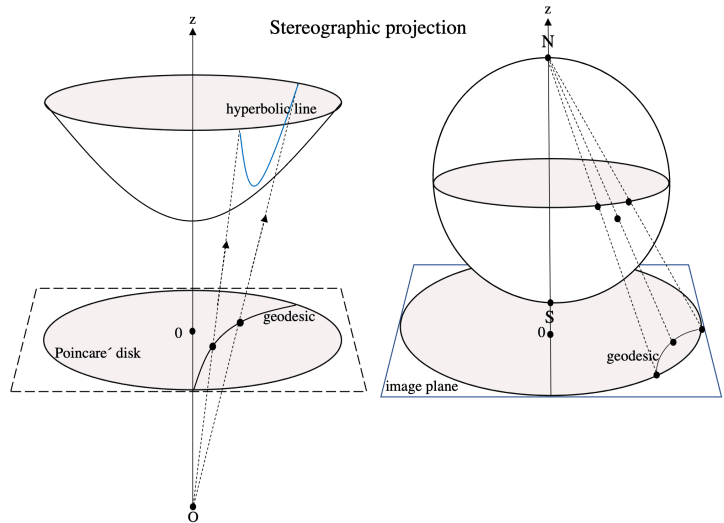}
   \caption{The stereographic projection of the second half sphere from north pole is analogue to the projection of hyperboloid on poincar\'e disk.}
\label{fig:stereo}
\end{figure}

 Convolutional neural networks (CNNs) have achieved state-of-the-art results on perception tasks when trained on perspective images. However, their performance significantly drops when applied on or transferred to fisheye images. Convolution networks rely on the translation invariant property and use fixed kernel shapes over all image plane. The translation invariance property makes their training harder on fisheye images because of the spatially varying distortion. Therefore, some CNN model adaptation methods based on kernel adaptation and learning were proposed to solve recognition tasks from large FoV images.   
Our work is inspired by the deformable kernel learning concept \cite{8237351, 8099683} which aims to learn local offset fields prior to convolution units in standard CNNs. Here, we propose a novel method that learns deformable kernels in hyperbolic spaces for wide FoV images; we generalize the original concept to provide a flexible solution for non-linear offsets. \frenchspacing
%%%%%%%%%%%%%%
\begin{figure}[t]
\centering
%\fbox{\rule{0pt}{2in} \rule{0.9\linewidth}{0pt}}
 \includegraphics[width=0.9\linewidth]{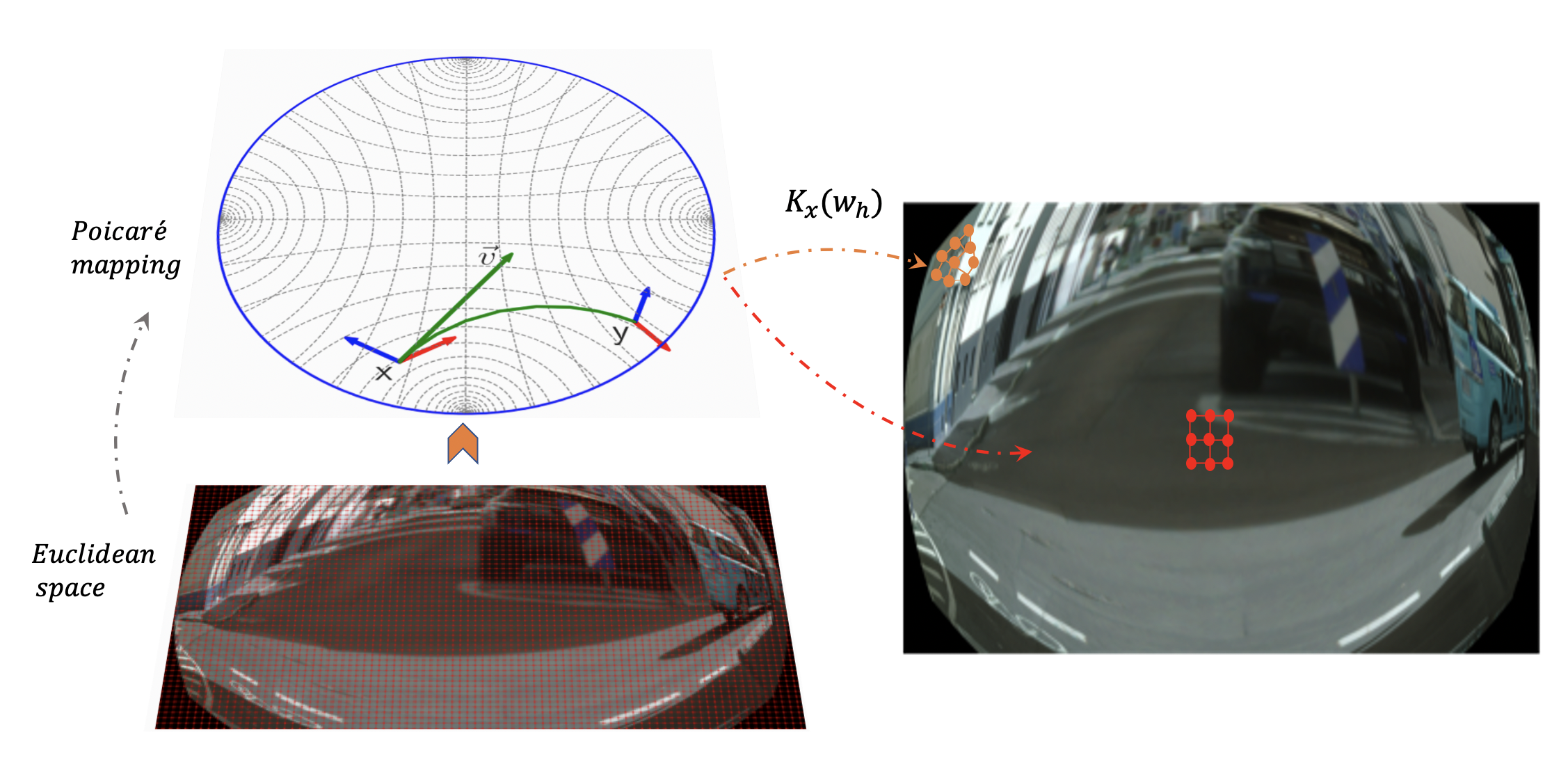}
   \caption{Learning deformable kernels in hyperbolic space for adapting CNNs to large FoV (fisheye) images. Image and input feature maps are represented on a graph and mapped to the poincar\'e disk for learning positions in a ($k\times k$) receptive field $K_{x}$ at every spatial location $x$.}
\label{fig:HyperCNN}
\end{figure}
%%%%%%%%%%%%%%
\paragraph{Motivation}
We are motivated by learning CNNs aware to the geometric distortions tied to wide FOV cameras such as fisheye.
We assume an existing analogy between hyperbolic space, more particularly the Poincar\'e Ball Model, and fisheye projections. The poincar\'e ball is a stereographic projection of the hyperbolic space on a disk in image plane (Fig. \ref{fig:stereo}). 
The poincar\'e ball is a conformal mapping that preserves angles between distorted lines. We can thus obtain a one to one correspondance (bijective mapping) between Euclidean and hyperbolic spaces for kernel sampling. Our intuition is to learn local displacements of deformable kernels on diffeomorphisme using geodesic (Poincar\'e) metric of non-Euclidean space to better capture smooth non-linear distortions of fisheye effects. 
\paragraph{Contributions} In this work, we introduce \emph{FisheyeHDK}, a hybrid neural network that combines hyperbolic and Euclidean convolution layers to learn the shape and weights of deformable kernels (Fig. \ref{fig:HyperCNN}), respectively. We build deformable kernel functions on top of existing convolution layers of Euclidean CNNs.  
Using synthetic fisheye data, we conduct extensive experiments and ablation studies to illuminate the intuition behind our approach and demonstrate its effectiveness. The data generation process implies converting narrow FoV datasets (perspective images) to distorted ones using fisheye polynomial model controlled by setting the focal length-like parameter at arbitrary values generating different severity levels of distortion.
For this task, we selected Cityscapes \footnote{\url{https://www.cityscapes-dataset.com/}} and BDD100K \footnote{\url{https://bdd-data.berkeley.edu/}} datasets with pixel-level annotations. Through our experiments we provide in-depth analysis on the effect of learning deformable kernels in both Euclidean and hyperbolic spaces. We show that our method improves the performance of CNN semantic segmentation by an average gain of 2\% on synthetic distortions. We also provide experimental results on a set of real-world images collected from fisheye camera and show that our method has better accuracy than baseline methods (more than 3\%). 
%%%%%%%%%%%%%%%
%
%%%%%%%%%%%%%%%

\section{Related works}  \label{sec:RW}
\paragraph{Fisheye augmentation} Recent research works have started studying how to directly adapt existing state-of-the-art object recognition models to fisheye images. In the context of object detection, Goodarzi \emph{et al.} \cite{8906325} have optimized a standard CNN detector for fisheye cameras through data augmentation in which synthetic fisheye effect was generated on training data using radial distortion. Ye \emph{et al.} \cite{ye2020universal} have recently learned a universal semantic segmentation CNN on urban driving images using a seven degrees of freedom geometric transformation as fisheye augmentation. Although, fisheye augmentation and fine-tuning are the simplest model adaptation techniques, there exist a fundamental limitation in applying these techniques with CNNs. The spatially variant (non-linear) distortion caused by large FoVs breaks the translation invariance property of the standard CNNs designed on regular grid structures \cite{Yogamani_2019_ICCV}. 
\paragraph{Kernels adaptation} Some works proposed to transfer planar CNNs on sphere assuming that fisheye lens produces spherical images  \cite{8906326, 8953831}. To minimize distortion, SphereNet \cite{8906326} adapts the sampling locations of convolution kernels using projection on sphere and tangent plane. Kernel Transformer Network \cite{8953831} uses equirectangular projection of spherical images to adapt the shape of the kernel based on matching activation maps between perspective and equirectangular images. These methods originally proposed for FoV cameras of 360$^{\circ}$ in which the resulted image can be perfectly projected on a sphere. The problem with fisheye cameras is that the FoV can vary between 180$^\circ$ and 280$^\circ$. This makes the sphere adaptation not an optimal solution for fisheye images as reported by \cite{Yogamani_2019_ICCV}. Fisheye image could be the result of one of four projections: Stereographic, Equidistant, Equi-solid and Orthogonal \cite{kannala06} based on the large FoV lens properties. An inverse mapping to a region on the sphere requires knowing a priori the FoV and center of distortion.  
%%%%%%%%%%%%%%
\paragraph{Learning deformable kernels}  Instead, learning deformable kernels is a projection-free concept which is generic and applicable to FoVs smaller than 360$^{\circ}$. Our work is related to this line of research.
Deformable Convolution Network (DCN) is originally proposed by \cite{8237351} and applied on perspective images to improve recognition tasks. The DCN augments standard convolution layers with learnable 2D offsets to sample deformable kernels for convolution operations. Here, kernels are sampled at each location in the spatial support of input features. Later, Playout \emph{et al.} \cite{palyout21} and Deng \emph{et al.} \cite{8842620} have directly applied this idea on fisheye images for semantic segmentation. However, to preserve the spatial correspondence between input images and the predicted semantic maps, Deng \emph{et al.} \cite{8842620} proposed a slight modification. They restricted offsets learning only in the neighbour locations of the kernel's center which kept unlearnable. This modified version of deformable convolution was called RDC \cite{8842620}. We argue that fixing the center of the kernel cannot resolve the fundamental limitation inherit in learning offsets in Euclidean space for wide FoV images.  
Compared to them, our method is more generic and flexible. It learns those kernels in hyperbolic space, more particularly, Poincar\'e ball model of hyperbolic space, and show that this model better captures fisheye effect than Euclidean methods. A new insight that could inspire future research on non-perspective cameras. The shape of convolution kernels is learnable and change flexibly over the spatial support, which makes CNN model adaptation applicable on large to ultra-wide FoV cameras. We provide a figure illustrating these effects on a toy example in supplementary material. 
% %-------------------------------------------------------------------------
\section{Review of hyperbolic geometry and the Poincar\'e Ball}
An $d$-dimensional hyperbolic space, denoted $\mathbb{H}^{d}$, is a homogenous, simply connected, $n$-dimensional Riemannian manifold with a constant negative curvature $c$. Analogous to sphere space (which has constant positive curvature), hyperbolic space is a space equipped with non-Euclidean (hyperbolic) geometry in which distances are defined by \emph{geodesics} (shortest path between two points). Hyperbolic space has five isometric models: the Klein model, the hyperboloid model, the Poincar\'e half space model and the Poincar\'e ball model \cite{Cannon97hyperbolicgeometry}. A mapping between any two of these models preserves all the geometric properties of the space. In this work, we choose the Poincar\'e ball model due to its conformal mapping properties (w.r.t. Euclidean space) and analogy to fisheye distortion. % or due to their isomorphic property%
The Poincar\'e ball model is defined by the Riemannian manifold ($\mathbb{D}^{d}_{c}$, $g^{\mathbb{D}_{c}}$), where $\mathbb{D}^{d}_{c} := \{x\in\mathbb{R}^{d} | \|x\| < 1/\sqrt{c}\}$ is an open ball of radius $1/\sqrt{c}$ and its Riemannian metric is given by $g^{\mathbb{D}_{c}}_{x} = (\lambda^{c}_{x})^{2}g^{\mathbb{E}}$ such that $\lambda^{c}_{x} := \frac{2}{1-c\|x\|^{2}}$ and $g^{E}=\mathbf{I}_{d}$ denotes the Euclidean metric tensor (the dot product). The induced distance between two points $x, y \in\mathbb{D}^{d}_{c}$ is given by
\begin{equation}
d_{\mathbb{D}^{c}}(x,y) = \frac{1}{\sqrt{c}} \cosh^{-1} \left(1+2\frac{\|x-y\|^{2}}{(1-c\|x\|^{2})(1-c\|y\|^{2})} \right)
\end{equation}
In hyperbolic space, the natural mathematical operations between vectors, such as vectors addition, subtraction and scalar multiplication are described with M\"obius operations \cite{ungar_gyrovector_2008}. The M\"obius addition of $x$ and $y$ in $\mathbb{D}^{d}_{c}$ is defined as
\begin{equation}
x\oplus_{c}y := \frac{(1+2c\langle x,y \rangle + c\|y\|^{2})x + (1-c\|x\|^{2})y} {1+2c\langle x,y \rangle + c^{2}\|x\|^{2}\|y\|^{2}}
\label{eq:add}
\end{equation}
and the M\"obius scalar multiplication of $x\in\mathbb{D}^{d}_{c} \setminus \{ \mathbf{0} \}$, $c>0$, by $a\in\mathbb{R}$ is defined as
\begin{equation}
a\otimes_{c}x := \frac{1}{\sqrt{c}} \tanh \left(a\tanh^{-1} \sqrt{c}\|x\|\right)\frac{x}{\|x\|}
\label{eq:mult}
\end{equation}
Note that subtraction can be obtained by $x\oplus_{c}(-1\otimes_{c}y) = x\oplus_{c}-y$. When c goes to zero, one recovers the natural Euclidean operations.
The bijective mapping between the Riemannian manifold of the Poincar\'e ball ($\mathbb{D}^{d}_{c}$) and its tangent space (Euclidean vectors $\mathcal{T}_{x}\mathbb{D}^{d}_{c}\cong \mathbb{R}^{d}$) at given point is defined by the exponential and logarithmic maps. To do so, Ganea \emph{et al.} \cite{ganea_hyperbolic_2018} derived a closed-form of the exponential map $\exp^{c}_{x} :\mathcal{T}_{x}\mathbb{D}^{d}_{c} \rightarrow \mathbb{D}^{d}_{c}$ and its inverse $\log^{c}_{x}: \mathbb{D}^{d}_{c} \rightarrow \mathcal{T}_{x}\mathbb{D}^{d}_{c}$ for $v\neq 0$ and $y\neq x$ such that
\begin{eqnarray}
\exp^{c}_{x}(v) =x\oplus_{c}\left( \tanh\left( \sqrt{c}\frac{\lambda^{c}_{x}\|v\|}{2} \right) \frac{v}{\sqrt{c}\|v\|} \right), \label{eq:exp} \\
\log^{c}_{x}(y)= \frac{2}{\sqrt{c}\lambda^{c}_{x}} \tanh^{-1}\left(\sqrt{c} \|-x\oplus_{c} y\| \right) \frac{-x\oplus_{c} y}{\|-x\oplus_{c} y\|} \label{eq:log}
\end{eqnarray}
As reported by \cite{ganea_hyperbolic_2018}, the maps have nicer forms when $x = \mathbf{0}$. This makes the mapping between Euclidean and hyperbolic spaces obtained by $\exp^{c}_{\mathbf{0}}$ and $\log^{c}_{\mathbf{0}}$ more useful in practical point of view.   
% %%%%%%%%%%%%%%%%
\begin{figure*}[t]
\centering
%\fbox{\rule{0pt}{2in} \rule{0.9\linewidth}{0pt}}
 \includegraphics[width=\linewidth]{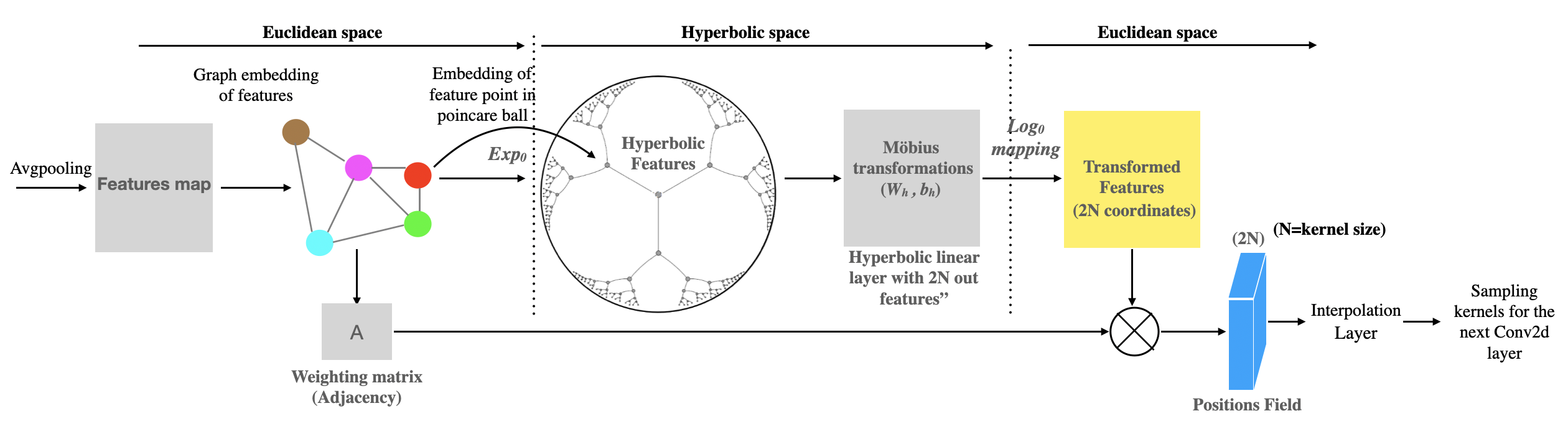}
   \caption{FisheyeHDK convolution layer. HDK network is one hyperbolic convolution layer predicting the positions in a deformable kernel of size ($k\times k$) for each point the spatial support. Conventional (Euclidean) convolution is applied between training weights sampled at predicted positions and input feature map.}
\label{fig:HDK}
\end{figure*}
%%%%%%%%%
\section{Proposed approach}
In this section, we introduce the proposed \emph{FisheyeHDK} approach. Our method does not require a ground truth information of fisheye geometry but updates the parameters of deformable kernels from optimizing CNN features through back-propagation. During training, hyperbolic deformable kernels are mapped to the Euclidean space and used in CNN layers to compute the output feature map (see Fig. \ref{fig:HDK}). Section \ref{sec:graph} explains how the input features are encoded in hyperbolic space to implement the HDK network. Section \ref{sec:net} presents the architecture of the HDK network and CNN implementation with deformable kernels.
% A final revision to see if I keep memory and comuptation and instead talk about downsampling and upsampling...instead of superpixel downsample the grid to a small size and talk about upsampling the output to recover the initial spatial dimensions
\subsection{Input representations for HDK network}  \label{sec:graph}
To leverage spatial information with feature vectors in hyperbolic space we represent input feature maps as graphs.
Images can naturally be modeled as a graph. The simplest graph model is defined on regular grid, where vertices correspond to pixels encoding features information and edges represent their spatial relations. This representation, however, requires considerable computations and memory for large grids. To alleviate such complexity and reduce inputs dimensionality, we downsampled the resolution of spatial grids by a factor of $2^{m}$ ($m=2$ as default). This allowed faster computations with insignificant effect on the performance, and enabled generating the graph from input features online.
%building graphs using the super-pixel approach \cite{6205760}. 
%It is a simple and fast approach that can convert grid data to graph online using CUDA implementations. 
We used CUDA implementations and the open source library Pytorch-geometric \footnote{\url{https://github.com/rusty1s/pytorch_geometric}} to generate graphs from grid feature maps.
%In all our experiments, we fixed the number of vertices (super-pixels) required to generate the graph at $70\%$ of the total size of the spatial grid. 
Consequently, the input to the HDK network are: vertices matrix $V\in \mathbb{R}^{N\times d}$, where $N$ is the number of vertices and $d$ is the feature dimension, and adjacency matrix $A$ of size ${N \times N}$ encoding the spatial information. Please refer to supplementary material for more details about image to graph generation.
%%%%%%%%%%%%%%%%%
%%%%%%%%%%%
%%%%%%%%%%%%%%%%%
\subsection{FisheyeHDK Architecture} \label{sec:net}
To implement FisheyeHDK, we build a hybrid architecture combining non-Euclidean (hyperbolic) convolution layers for kernel's positions learning and Euclidean convolution layers for features learning. Euclidean convolution layers belong to a conventional CNN model chosen from existing architectures. These layers apply convolution on an input feature map using deformable kernels sampled from the output of hyperbolic layers as shown in Figure~\ref{fig:HDK}. 
Hyperbolic convolution layers use preceding feature maps converted to graph to learn the shape of deformable kernels in hyperbolic space. We describe in the following the hybrid components of the proposed architecture. 
\paragraph{HDK network} comprises one hyperbolic convolution layer. Euclidean feature vectors are projected on hyperbolic space using Exp. map as described in Eq. (\ref{eq:feat}): 
\begin{equation}
H_{v} = \exp^{c}_{\mathbf{0}}(F_{v}),
\label{eq:feat}
\end{equation}
where $F_{v}$ is the Euclidean feature vector, and $H_{v}$ is its projection on hyperbolic space for a give vertex $v$.
A M\"obius layer performs linear transformations on feature vectors inside Poincar\'e Ball (Eqs. (\ref{eq:add}) and (\ref{eq:mult})). M\"obius features are mapped to the Euclidean space using the Log mapping as shown in Fig.~\ref{fig:HDK}, and the spatial information encoded by the adjacency matrix are aggregated with projected features on the tangent (Euclidean) space using an aggregation layer such as
\begin{equation}
K = \log^{c}_{\mathbf{0}}\left(\left(W_{h}\otimes_{c} H \right)\oplus_{c} b_{h} \right)\odot A,
\label{eq:agg}
\end{equation}
where $\odot $ denotes element-wise product, $W_{h}$ and $b_{h}$ are hyperbolic weights and bias vectors, $K$ is a dense map of deformable kernels representing the positions inside ($k\times k$) window at every point of the grid. We used the open source library geoopt \footnote{\url{https://github.com/geoopt/geoopt}} to build the HDK network. Additional details about the HDK network are provided in supplementary material.
%%%%%%%%%%%
\section{Experiments}
In this section, we present our experiments on the task of semantic segmentation. For the sake of generalization to other recognition tasks, we focus on the encoder components of the CNN model and demonstrate the effectiveness of hyperbolic deformable kernels versus their Euclidean counterparts for large FoV images. We tested our approach on three datasets. Two datasets of perspective images were transformed to fisheye using the general fisheye model. We rely on the model used in the open source library OpenCV \footnote {\url{https://docs.opencv.org/3.4/db/d58/group__calib3d__fisheye.html}}. The third dataset are real-world images collected by a fisheye camera.
%%%%%%%%%%%%%%
\begin{table*}[t]
\centering
\begin{tabular}{l|ccc|ccc|c}
\hline
\multirow{2}{*}{Method}	     &           &  $\#$ First layers ($l$)      &            &           & $\#$ Last layers ($l$)  &            &  \\ \cline{2-8} 

             & $l=1$ & $l=3$                         & $l=6$ & $l=1$ & $l=3$                     & $l=6$ & $l=all$ \\
\hline\hline
RDC \cite{8842620} & 57.9 & 58.0& 57.9& 57.8& 57.7& 57.9&56.8\\
FisheyeHDK (Ours)& 58.3& 58.4& 58.5& 58.5 & 59.6& 59.3 & 58.0 \\
\hline
\end{tabular}
\caption{Test results of Mean intersection over union (mIoU (\%)) on distorted cityscapes with $f=200$. Deformable kernels are applied on ResNet101 module. First and last layers are convolutions starting from $\text{conv1}$ and $\text{layer4.1.conv3}$, respectively.}
\label{tbl:f200}
\end{table*}
%%%%%%%%%%%%%%
\subsection{Datasets} 
\paragraph{Cityscapes} are perspective images with pixel-level annotations collected from urban German cities. The dataset comprises 5000 images divided into train, validation and test sets (2975, 500 and 1525 images respectively). Annotations consist of 30 classes but only 19 are defined as valid. The test set is provided without annotation maps, therefore we only used train and validation sets.  The validation set was used as the test set and the original training set was split in two (0.9/0.1 ratio) for training and validation purposes. Their original resolution is ($1024\times 2048$) pixels. Three fisheye datasets were generated from perspective cityscapes images using the following focal lengths: 200, 125 and 50. Images and ground truth maps are resized to ($512\times 1024$) pixels in all three datasets.
\paragraph{BDD100K} A large-scale divers dataset recently released for perception tasks in autonomous driving. It was collected from divers and complex environments in US cities. The dataset for the segmentation task is composed of 10000 perspective images with fine-grained pixel-level annotations (including 40 object classes). Images and annotations are divided into train, validation and test sets (7000, 1000, 2000, respectively). Their original resolution is ($720 \times 1280$) pixels. Annotations are not provided in the test set. Similar to Cityscapes, we use the validation set as test set and split the original train set into two sets: 6500 for training and 500 for validation. For the experiments conducted on this dataset, we generated two fisheye datasets using smaller focal lengths set to the values of 75 and 50. Images and ground truth maps are resized to ($700\times 1024$) pixels.
%%%%%%%%%%
\paragraph{Real Data} A dataset collected using real fisheye cameras mounted in a driving vehicle. The dataset comprises 800 real-world images with pixel-level annotations similar to Cityscapes. Annotations include few set of classes which are: person, car, train, truck, traffic light, cyclist, motorcycle and bus. The resolution of fisheye images is ($512 \times 1024$) pixels. We randomly split the dataset into three sets: 680 for training, 20 for validation and 100 images for testing. We kept the original resolution of images during all training phases. We apply data augmentation by randomly left-right flipping and by randomly changing color information (brightness, contrast, hue and saturation) during training.
%%%%%%%%%%%%%%%%%%%%%%%%%%%%%
\subsection{Implementation details} \label{sec:imp}
\paragraph{Training protocol} Without loss of generality, we built hyperbolic deformable kernel components on top of the DeepLabV3 architecture. Any alternative semantic segmentation network can work. 
We trained the model on 2 GPUs using synchronized batch-norm. We implemented the pixel-wise weighted cross-entropy as a loss function. For baselines, we used Stochastic Gradient Decent (SGD) optimizer with momentum $0.9$ and weight decay $5\times 10^{-4}$. The learning rate was initialized to $1\times 10^{-3}$ for the encoder and $1\times 10^{-2}$ for the decoder; both are updated using the "poly" learning rate policy. Our approach comprises both Euclidean and hyperbolic parameters. For hyperbolic parameters, we adopted the Riemannian SGD (RSGD) \cite{6487381} with a learning rate ($1\times 10^{-2}$) since they are manifold parameters and used usual SGD for the Euclidean parameters.
We provide more details on optimization in supplementary material. %
We initialized the encoder and decoder parameters of the CNN layers with ImageNet weights. We initialized the hyperbolic weights $W_{h}$ using the Xavier uniform distribution and hyperbolic biases $b_{h}$ with zeros for HDK layers. Euclidean deformable kernel weights were initialized by zeros as in \cite{8842620}. 
For the synthetic fisheye dataset, we set the training batch size to 16 and the validation batch size to 4. For real fisheye data, we set the batch size to 8 during training and validation. In all experiments, we trained the models for 100 epochs. We used per-class accuracy and the standard mean Intersection-Over-Union (mIoU) as evaluation metrics for validation (after each epoch) and testing on test sets after training. We report evaluation metrics on the "valid" classes provided with Cityscapes and BDD100K datasets. Void class, corresponding to unsegmented or irrelevant objects, and image's borders resulted from fisheye transformation were ignored. 
%%%%%%%%%%%%%%%%%
%%%%%%%%%%%%%%%
\begin{table}[t]
\centering
\begin{tabular}{l|p{1.2cm}|p{1.2cm}|p{1.2cm}|p{1.0cm}}
\hline
\multirow{2}{*}{\shortstack[l]{Distortion \\ level (f)}} & \multicolumn{4}{c}{\shortstack[c]{\\Cityscapes Dataset \\ mIoU/mAcc (\%) \vspace{0.5mm}}} \\ \cline{2-5}
{} &\shortstack[c]{ \\Rect+Seg} & \shortstack[c]{ \\RegCNN} &\shortstack[c]{ \\RDC}  & \shortstack[c]{ \\Ours} \\
\hline\hline
50  & 16.3/25.8& 45.8/61.2& 48.6/60.8 & \textbf{49.8}/\textbf{61.5}  \\
125 & 44.6/51.9& 51.3/70.0& 54.3/69.4& \textbf{55.9}/\textbf{70.0}  \\
200 & 54.4/62.1& 53.6/71.8& 57.7/71.6& \textbf{59.6}/\textbf{72.1}  \\
\hline
{} & \multicolumn{4}{c}{\shortstack[c]{\\\vspace{0.5mm} BDD100K Dataset \\ mIoU/mAcc (\%) \vspace{0.5mm} }} \\ \cline{2-5}
50 & 6.9/12.3& 30.2/40.6& 40.2/52.7& \textbf{40.5}/\textbf{53.0}\\
75 & 16.9/24.7 & 38.9/51.7& 42.8/\textbf{57.6}& \textbf{44.0}/56.6 \\
\hline
\end{tabular}
\caption{Effect of distortion on the performance of segmentation model augmented with deformable kernels in last 3 layers of ResNet101. Smaller is $f$, stronger is the distortion. We report the results on the test sets of dataset transformed to fisheye.}
\label{tbl:fl}
\end{table}
%%%%%%%%%%%%%
\begin{table}[t]
\centering
\begin{tabular}{l|p{1.5cm}|p{1.2cm}|p{0.8cm}}
Method & train/val speeds (on epoch) & test time (on one image) & Size (MB) \\
\hline
RegCNN & \textbf{634.3}/\textbf{33}s & 0.270s & \textbf{237.9} \\
RDC \cite{8842620} & 656/34s & 0.278s& 238.9\\
\hline
FisheyeHDK-4 & 680/44s & 0.318s & 238.0 \\
FisheyeHDK-8 & 650/35s & 0.288s&238.0 \\
\hline
\end{tabular}
\caption{Comparative results using cityscapes dataset (2677 train images, 298 val images) converted to fisheye. All networks are trained on 2 Nvidia Tesla P100 GPUs each with 16Gb memory. Images resolutions is ($512\times 1024$) pixels}
\label{tbl:speed}
\end{table} 
%%%%%%%%
%%%%%%%%%%%%%%%
\begin{table*}[t]
\centering
\begin{tabular}{l|cccccccc}
\hline
\multirow{2}{*}{Method}	     &   \multicolumn{8}{c}{Per-Class IoU / Per-Class Accuracy (\%)} \\ \cline{2-8} 

{}             & traffic light & person  & car & truck & bus  & train & motorcycle & mIoU/mAcc \\
\hline\hline
Rect+Seg & - & 26.7/29.2& 57.5/70.6 & - & - & - & -  & 42.5/49.9\\
RegCNN & 65.6/88.6 & 86.7/88.3 & 77.8/93.4 & 72.7/79.7 & 58.6/1.00 & 25.8/26.3 & 69.7/69.8 & 65.3/78.0 \\
RDC \cite{8842620} & \textbf{64.5}/83.6 & \textbf{96.5}/\textbf{97.8} &78.9/94.0 & 69.4/75.6 & \textbf{71.8}/99.0 & 79.4/\textbf{85.9} & 73.3/73.8 & 76.3/87.2 \\
FisheyeHDK & 63.0/\textbf{91.5} & 96.2/97.4 & \textbf{83.6/95.2} & \textbf{81.7/85.4} & 71.7/99.0 &\textbf{81.9}/83.2& \textbf{82.9/83.1} & \textbf{80.1/90.8} \\
\hline
\end{tabular}
\caption{Per-class accuracy and IoU (\%) on the test set of real fisheye images.}
\label{tbl:acc_real}
\end{table*}
%%%%%%%%%%%%%%%%%%
\subsection{Baseline models}
We compare our approach with conventional CNN models, and with existing deformable convolution methods \cite{8237351, 8842620}. We choose one as both learn offsets in same way and our preliminary tests did not show significant differences (see supplementary material for details). We report in this paper the experiments conducted on the RDC method \cite{8842620}. Even though deformable convolutions can be built on any state-of-the-art CNN architecture, we limit our implementations to one CNN semantic segmentation model and focus on the effect of hyperbolic and Euclidean deformable kernels.   
We re-implement the RDC approach on fisheye datasets that we generated with different focal length values using same training protocols. 
As a baseline CNN segmentation model, we choose the DeepLabV3 architecture in which deformable kernels are applied in the backbone component of the encoder (ResNet101). We report additional results on other backbone architectures in supplementary material. Accordingly, two baseline models were used for comparisons: a regular CNN without deformable kernels (RegCNN) and a deformable one based on \cite{8842620}. Additional to these baselines, we compared with fisheye undistortion technique (based on inverted equidistant projection) prior to a standard pre-trained segmentation network (Rect+Seg). The segmentation network is a DeepLapV3 trained on original cityscapes images.
\subsection{Ablation study}
\begin{figure*}[t]
%\begin{center}
 \includegraphics[width=1.0\linewidth]{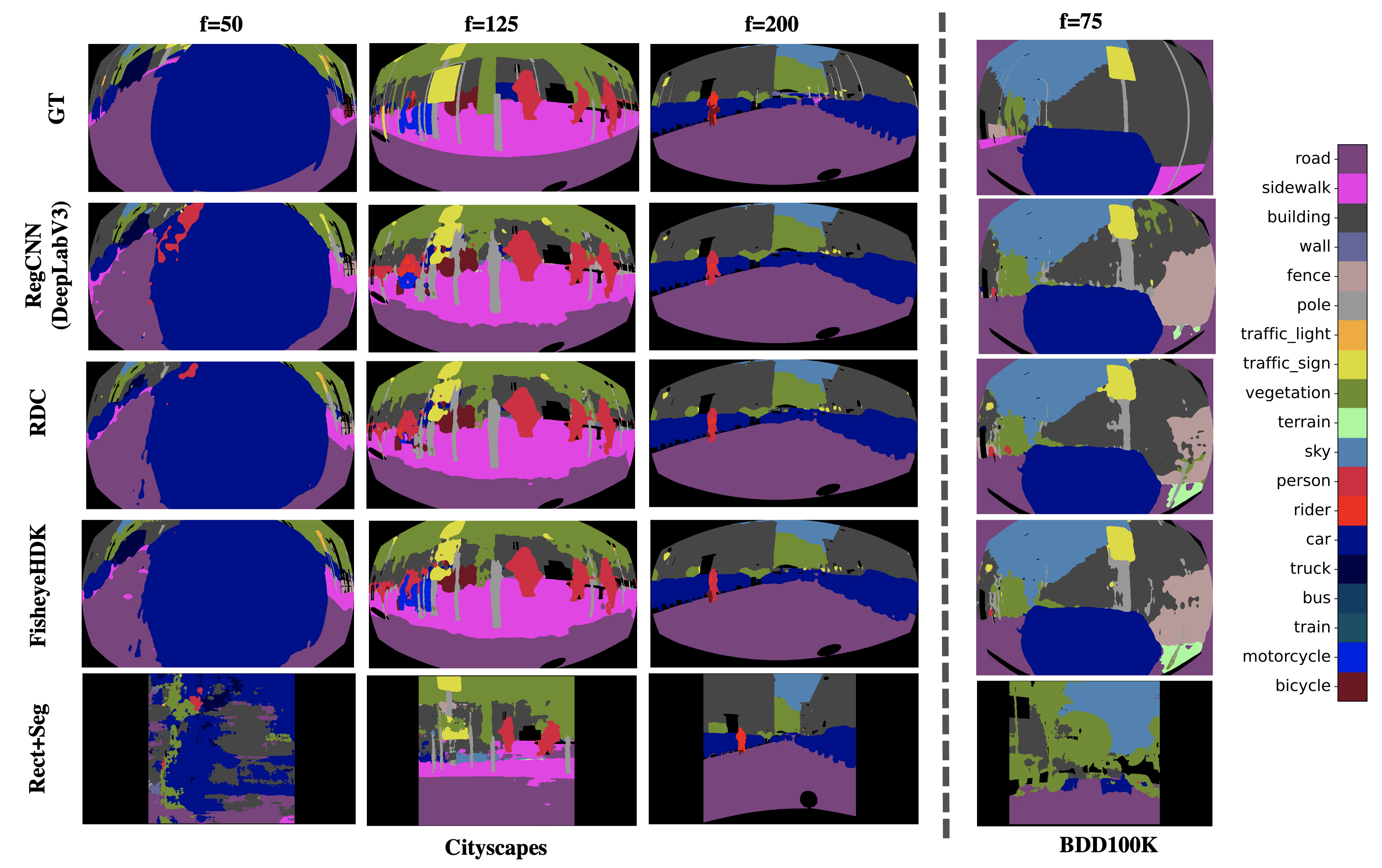}
 %\end{center}
 \caption{Qualitative results of different segmentation methods. On BDD100K dataset, we considered the road label as a void class.}
\label{fig:fmaps}
\end{figure*}

We investigate the effect of using deformable kernels in CNN for distorted images. Then, we analyze the distortion level's impact on the CNN model's performance with an emphasize on the segmentation task. Finally, we analyze the efficiency of our approach compared to baselines.  
\paragraph{Connection of Deformable Kernels and Distortion} % to be re-defined
%This study is split in two parts. First, we answer the question: \emph{\textbf{"which CNN layers needed to be deformable to learn better representations from distorted inputs?"}}.    
In this study, we answer the question: \emph{\textbf{"which CNN layers needed to be deformable to learn better representations from distorted inputs?"}}. 
We hypothesize that low-level and high-level features are strongly correlated with fisheye distortion. Low-level layers capture spatial information such as distorted edges and lines, while high-level features represent semantic information of distorted objects. Typically, distortion is more visible in low-level features, however, it is unclear to what extent deformable kernels used only on first layers would improve learning in the following convolution layers.  
In baseline methods \cite{8237351, 8842620}, convolutions with deformable kernels were limited to the few last layers of the CNN. To provide a deeper understanding on the relation between deformable kernels and fisheye distortion, we experiment all layers with emphasize on low-level and high-level layers. Therefore, both HDK and RDC networks were added prior to the CNN layers of the backbone module. We started with low-level layers and progressively added deformable kernels to $l$ convolutional layers, ($l=\{1, 3, 6\}$), then we did the same protocol on the last layers and finally on all layers. For this analysis, we used perspective cityscapes dataset transformed to fisheye with $f=200$. Table \ref{tbl:f200} shows the mIoU results of the segmentation model. 
As expected, augmenting low-level and high-level convolution layers with learned positions improves the performance better than using deformable kernels in all layers. According to these results, we can apply deformable convolution on few low-level or high-level layers and have almost similar performance. In the next of our experiments, we select as default the backbone variant with deformable convolution in the last three layers. Moreover, Table \ref{tbl:f200} shows that HDK networks improve the performance of the model compared to their Euclidean counterparts.
%%%%%%%%%%%%
\paragraph{Distortion effect on CNN performance}
The level of distortion in large FoV images depends on fisheye lenses. Severe geometric distortions result from using lenses with small focal length ($f$) to compensate wide FoV angles ($\theta$) when rays $(R)$ projected on finite image plane ($R = f.\theta$). In this experiment, we test our approach and the baseline methods on perspective datasets transformed to fisheye using different distortion levels: $f\in\{50, 75, 125, 200\}$. Table \ref{tbl:fl} reports the quantitative results of the performance of segmentation model. Figure \ref{fig:fmaps} shows qualitative results of different methods.
As expected regular CNN is the worst-performing on distorted images. Our approach improves regular CNN better than RDC approach on both datasets (Cityscapes and BDD100K). Undistortion has bad influence on segmentation performance and leads to a loss of FoV.% continue in details for which f and comment on images
%%%%%%%%
%%%%%%%%%
\paragraph{Efficiency and Model Size}
As mentioned in Section \ref{sec:graph}, HDK inputs are represented as structure of graph with $N$ nodes and $(N\times N)$ adjacency matrix. Feature maps should be converted to graph during training which adds additional computational cost to the model. The training speed decreases significantly when deformable kernels applied on high spatial resolutions. To keep efficiency comparable to regular CNN and RDC, we tested the performance versus training speed using two downsampling factors $2^{m}$ with $m=\left[2, 3\right]$. We refer to these networks by FisheyeHDK-4 and FisheyeHDK-8, respectively. Table \ref{tbl:speed} lists the comparative results with baseline methods. % 
Our method is efficient and more accurate than the deformable convolution network when spatial inputs are downsampled by a factor of 4. Our method becomes significantly slower when using lower downsampling ($m=1$). This is due to the computational cost needed to compute adjacency and upsample the kernels field up to the spatial resolution of the features input. Furthermore, strong downsampling leads to loss of information which could degrade the performance of our approach. Our approach does not significantly increase the size of the baseline CNN architecture compared to the RDC. As illustrated in Table \ref{tbl:speed}, all methods are close to the baseline CNN when deformable kernels applied on the last three layers. However, this result changes when deformable kernels networks are used prior to all backbone layers. Our model remains relatively light $238.5$MB (+0.25\%), while the model size of the of RDC method increases to $243.3$MB (+2.2\%). In supplementary, we provide a comparison between inference times of both methods.
%We show in Fig.~\ref{fig:time} the inference time on a single image at different configurations. 
%\begin{figure}
%\centering
% \includegraphics[width=0.8\columnwidth]{figures/inference_time.pdf}
 %  \caption{Inference time on single image ($512\times 1024$ pixels)}
%\label{fig:time}
%\end{figure}
%%%%%%%%
\subsection{Evaluation on real data}
Given the small size of this dataset, we applied transfer learning of feature weights from the baseline segmentation model trained on perspective cityscapes to real fisheye dataset and trained FisheyeHDK network for only 20 epochs. Deformable kernel network parameters were initialized as explained in Section \ref{sec:imp}, and used only in the last 3 layers of the backbone module. Table \ref{tbl:acc_real} shows the quantitative results of per-class metrics (accuracy and IoU).
The averaged metrics show that our method outperforms the baseline methods. On real fisheye distortion, the results prove that deformable kernel methods are better than using regular kernels in conventional CNN models. Segmentation after undistortion seems highly affected by the loss of FoV on this dataset. Some classes are ignored because they become less representative after undistortion. It is worth noting, for future work, that the small size of data reflects the reality of our world where the access and cost of annotating data is crucial. Further improvements and validations of our approach would consider this bottleneck.     
%--------------------------------

\section{Conclusion}
In this work, we introduced FisheyeHDK, a method that adapts regular CNN models on large FoV images based on deformable kernel learning. We proposed a novel approach that learns the shape of deformable kernels (positions) in hyperbolic space and demonstrated its effectiveness on synthetic and real fisheye datasets. For the first time, we empirically demonstrated that hyperbolic spaces could be a promising approach for deformable kernel sampling and CNN adaptation to ultra-wide FoV images. Our goal is to provide new insights and inspire future research on non-Euclidean (hyperbolic) geometry for learning deformations from ultra-wide FOV images. We do not intend, in this work, to providing high metrics analogue to perspective images instead to show that deformable kernels learned in Euclidean space are not the optimal solution for fisheye images. We keep further improvements to future work and we will explore the feasibility of using our approach in object detection models. An exciting line of research is to examine the capability of FisheyeHDK model to adapt faster to different tasks without retraining the Euclidean part (features) of the network. That could be crucial due to the lack of annotated fisheye data.

\section*{Acknowledgments}
The authors would like to acknowledge and thank Loic Messal from Jakarto (\url{https://www.jakarto.com}) for providing the private fisheye dataset and annotations. We also acknowledge Thales for funding this research work and the infrastructure team for their support on running our experiments on Kubernetes Platform. 

\bibliographystyle{unsrt}  
\bibliography{paperbib}

\end{document}

\maketitle

\section{Details of Image-to-Graph Generation}
To map images to a graph structure, we considered regular graph structure for all images and the feature maps of the intermediate layers where deformable kernels were applied. The graph's nodes are pixels locations and their corresponding feature vectors, the edges represent the spatial relation between pixels. 
The corresponding graph $\mathcal{G}= (V, E, \mathcal{A})$ is defined such that $V=(\mathbf{x}_{1}, ..., \mathbf{x}_{N})$ is the graph nodes ($\mathbf{x}_{i}$ is the pixel coordinates and $N$ is the number of node) and $E$ is the edges between connected pixels. Each node ($\mathbf{x}_{i}$) is characterized by feature vector of dimensions $d$, where $d$ represents the number of channels. The spatial relations between nodes of the graph are encoded by the adjacency matrix $\mathcal{A} = (a_{ij})$, where $a_{ij}= a_{ji}$, with $a_{ij} = 0$ if $(i,j)\not\in E$ and $a_{ij} = 1$ if $(i,j)\in E$. The resulted adjacency matrix is a sparse matrix of ($N\times N$) dimensions. We generate graphs from spatial inputs online during training for the set of convolutional layers chosen to operate with deformable kernels since the activation maps change during training. In the case of applying deformable kernels on the first convolution layer, which takes original images as input, the graph is generated only once prior to training. As we explain in section~\ref{sec:HDK}, we downsampled the features maps used in hyperbolic neural network layers for computational efficiency needs. We explain this in the following section. 

%%%%%%%%%%%%%%
\section{Toy example}
\begin{figure}[t]
\centering
%\fbox{\rule{0pt}{2in} \rule{0.9\linewidth}{0pt}}
 \includegraphics[width=0.8\linewidth]{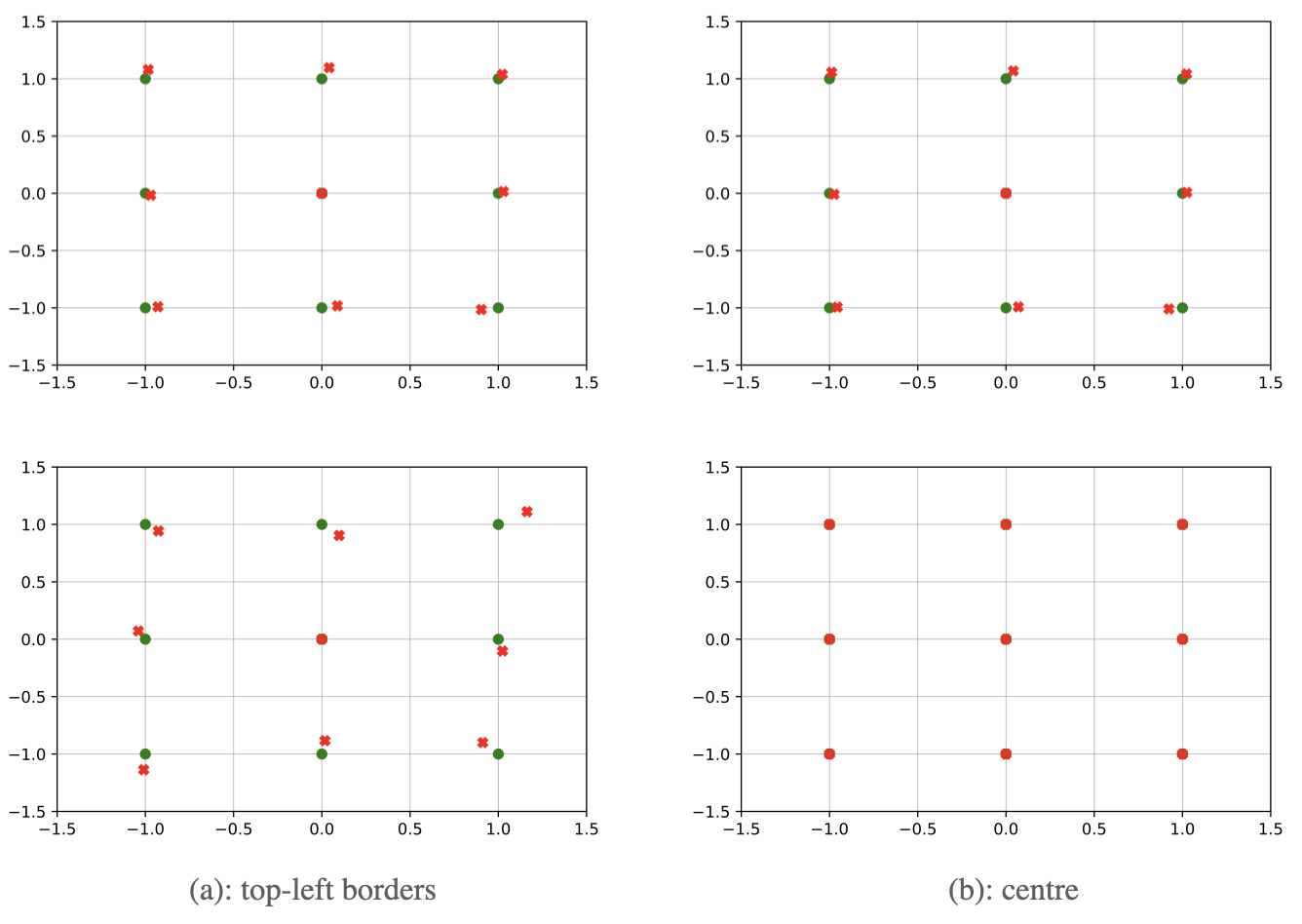}
   \caption{Predicted positions in a ($3\times 3$) kernel in Euclidean space (top) and hyperbolic space (bottom) after 20 epochs training on synthetically distorted images.}
\label{fig:kernels}
\end{figure}

In this example, we used the deepLabV3 network pertrained on cityscapes. We trained the network for 20 epochs on synthetically distorted images with focal length equals to 50. We used one RDC (restricted deformable convolution) and our HDK (each is composed of only one layer) to predict the offset fields prior to the first convolution layer of the encoder. We sampled two deformable kernels, one from the center and another from the top-left boundaries of a fisheye image. We show the kernels in Fig. \ref{fig:kernels}.  As can be observed, after 20 epochs of training, both methods predict non-zero offsets near the boundaries of the image. We can see that our method predicts larger offsets than RDC; this is due to the sever distortion level applied to the image. Most importantly, our method provides more accurate prediction of the kernel in image center where the distortion is zero. 
%%%%%%%%%%%%%%
%

%%%%%%%%%%%%%%%%%%%%%%%
\section{Details of FisheyeHDK Architecture}
We provide detailed information about our hybrid architecture comprising both hyperbolic and Euclidean layers. For synthetic datasets (Fisheye-Cityscapes and Fisheye-BDD100K), we train the model for 100 epochs on two Nvidia Tesla V100 GPUs with batch size of 8 images per GPU. Complete training took approximately 24 hours for cityscapes, and 30 hours for BDD100K images. We trained real datasets for 20 epochs since the dataset is relatively small. Complete training of this dataset took approximately 30 minutes. Note that we are restricted to a small dataset due to the cost of acquiring and annotating real data.
\subsection{HDK Network} \label{sec:HDK}
\begin{figure}[t]
\includegraphics[width=1.0\linewidth]{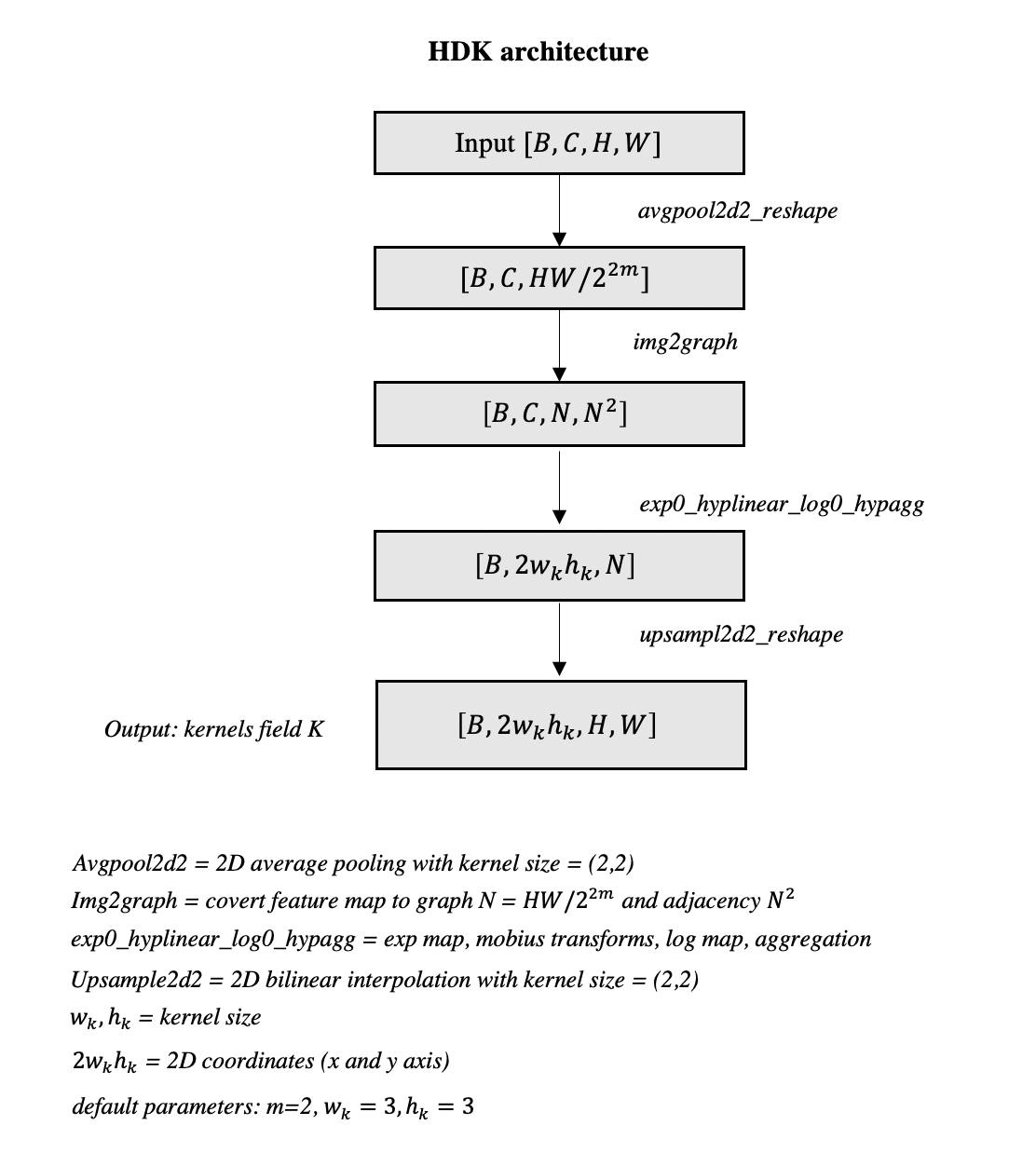}
\caption{Detailed HDK network architecture.}
\label{fig:HDK}
\end{figure}

 The adjacency matrix complexity is $O(N^{2})$ and could require expensive computations on feature maps with high spatial resolutions and slow down the training process. To ensure a reasonable trade-off between spatial resolution and computations for hyperbolic networks implemented on images, we added downsampling and upsampling layers.  The downsampling layer reduces the spatial resolution of the feature maps by averaging neighbourhood pixels with a factor of $2^m$. Accordingly, an input of shape $\left[B, C, H, W\right]$ with $B$, $C$, $H$, and $W$ denote batch size, channels, height and width of feature map respectively, the corresponding downsampled input becomes $\left[B, C, H/2^m, W/2^m\right]$. Then, the instances (feature maps) of the batch are reshaped to $\left[B, C, HW/2^{2m}\right]$ where the spatial positions are regarded as vertices of the $B$ graphs with number of vertices $N=HW/2^{2m}$ and feature dimension $d=C$. For example, if the value of $m$ is equal to $2$, then we get $N=HW/16$.\\ 
The first layer of the HDK architecture maps the feature vector $F_{v}$, at vertex $v$, to the hyperbolic space as defined in the main paper. The second layer performs  linear regression and learns kernel shape embeddings on the poincar\'e ball (xavier initialization is used for the hyperbolic weights $W_{h}$, while hyperbolic biases $b_{h}$ are initialized with zeros). The third layer re-projects the embeddings in the tangent space and aggregates neighbour's embeddings of center node using the binary adjacency matrix $\mathcal{A}$ as shown in the main paper. 
% The adjacency matrix allows to compute a weighted average of the positions 
%Inspired by the graph convolution neural networks, the aggregation operation is a weighted average of neighbours embeddings of the center node (position). It allows propagating information over neighbourhood positions such that position coordinates of every node are calibrated by the information encoded at their neighbour positions. 
The last layer performs upsampling with bilinear interpolation to reconstruct the spatial resolution of the kernel's field which must be aligned with the spatial resolution of the feature map ($\left[B, C, H, W\right]$) before downsampling. This layer outputs the displacement field which defines the predicted positions inside every $h_{k}\times w_{k}$ kernel at each position in image. Hence, the shape of the output is $\left[B, 2h_{k}w_{k}, H, W\right]$ where each spatial position of the features maps in the batch has $2h_{k}w_{k}$ values indicating 2D pixel coordinates. Figure \ref{fig:HDK} shows the operations details of the HDK architecture. It is worth noting that the adjacency matrix becomes intractable for high resolution graphs which justifies our choice of adding downsampling and upsampling layers to the HDK architecture. 
%%%%%%%%%%%%%%%%%%%%%%%%%%

\subsection{Optimization}
Both CNN and HDK network components are optimized simultaneously. The SGD optimizer is used to update the convolution parameters $\theta_{e}$ of the CNN, while optimization with respect to hyperbolic parameters $\theta_{h}=(W_{h}, b_{h})$ is based on Riemannain gradients which are the rescaled Euclidean gradients. Thanks to the conformal mapping between Poincar\'e model and Euclidean space, back-propagation runs in the standard way as reported by \cite{nickel_poincare_2017}: 
\begin{equation}
\theta^{t+1}_{h} = \theta^{t}_{h} - \eta^{t} (1-\|\theta^{t}_{h}\|^{2})^{2}/4 \nabla_{E}
\end{equation}
where $\nabla_{E}$ is the Euclidean gradient of the hyperbolic layer w.r.t. to the first-order Taylor approximation. In our case, it is the Euclidean gradient of the HDK layer's outputs (Eq. (3) in the main paper): 
\begin{equation}
\nabla_{E} = \partial K/\partial \theta^{t}_{h} \approx (K(\theta^{t+1}_{h})-K(\theta^{t}_{h}))
\end{equation}
Accordingly, two parameters updates were performed simultaneously at each training step.
%%%%%%%%%%%%%%%%%%%%%%%%%%
\section{More Ablation}
In the main paper, we focused on our main idea about learning deformable kernels in hyperbolic space (particularly the Poincar\'e Ball model) for semantic segmentation of large FOV images obtained by fisheye cameras. We used DeepLab model with ResNet-101 architecture as backbone of its encoder component. For the sake of generality, we provide here additional quantitative results of our ablations performed on different architectures which have been widely used with DeepLab model. The results can be found in Tables \ref{tbl:layers_abl} and \ref{tbl:ablation}. 
\begin{table*}[t]
\begin{center}
\begin{tabular}{l|cc|cc} %{l |c|c|c}
\multirow{2}{*}{Backbone} &   $\#$ First layers ($l$)      &         & $\#$ Last layers ($l$)   \\ \cline{2-5}   \\
& $l=1$ & $l=3$ & $l=1$ & $l=3$  \\
\hline\hline
ResNet-50& 56.6 & 56.9& 56.9 & 56.8\\
 MobileNet-V2 & 52.2 & 52.3 & 52.4& 54.3\\
 DRN &  61.9 & 62.3& 63.3 & 64.2\\
 \hline
\end{tabular}
\end{center}
\caption{Results of mean intersection over union (mIoU (\%)) on distorted cityscapes test set with $f=200$ using different backbone modules.}
\label{tbl:layers_abl}
\end{table*}
\begin{table}[t]
\begin{center}
\begin{tabular}{p{2.5cm}|p{1.0cm}|p{1.0cm}|p{1.0cm}}
\hline
 \shortstack[l]{\\backbone\\ architecture} & f=50 & f=125 & f=200 \\
\hline\hline
ResNet-101 & 49.8 & 55.9 & 59.6 \\
ResNet-50 & 47.1 & 54.4 & 56.8 \\
DRN & 54.2 & 63.7 & 64.2 \\
MobileNet-V2 & 42.3 & 52.1 & 54.3\\
\hline
\end{tabular}
\end{center}
\caption{Quantitative results of our ablation on cityscape dataset distorted with f=50 (higher distortion), 125 and 200. Deformable kernels were applied on the last convolution layers ($l=3$) of each backbone architecture.}
\label{tbl:ablation}
\end{table}
%%%%%%%%%%%%%%%%%%%%%%%%%%%%%
\section{More comparisons}
\begin{figure}
\centering
\includegraphics[width=0.8\columnwidth]{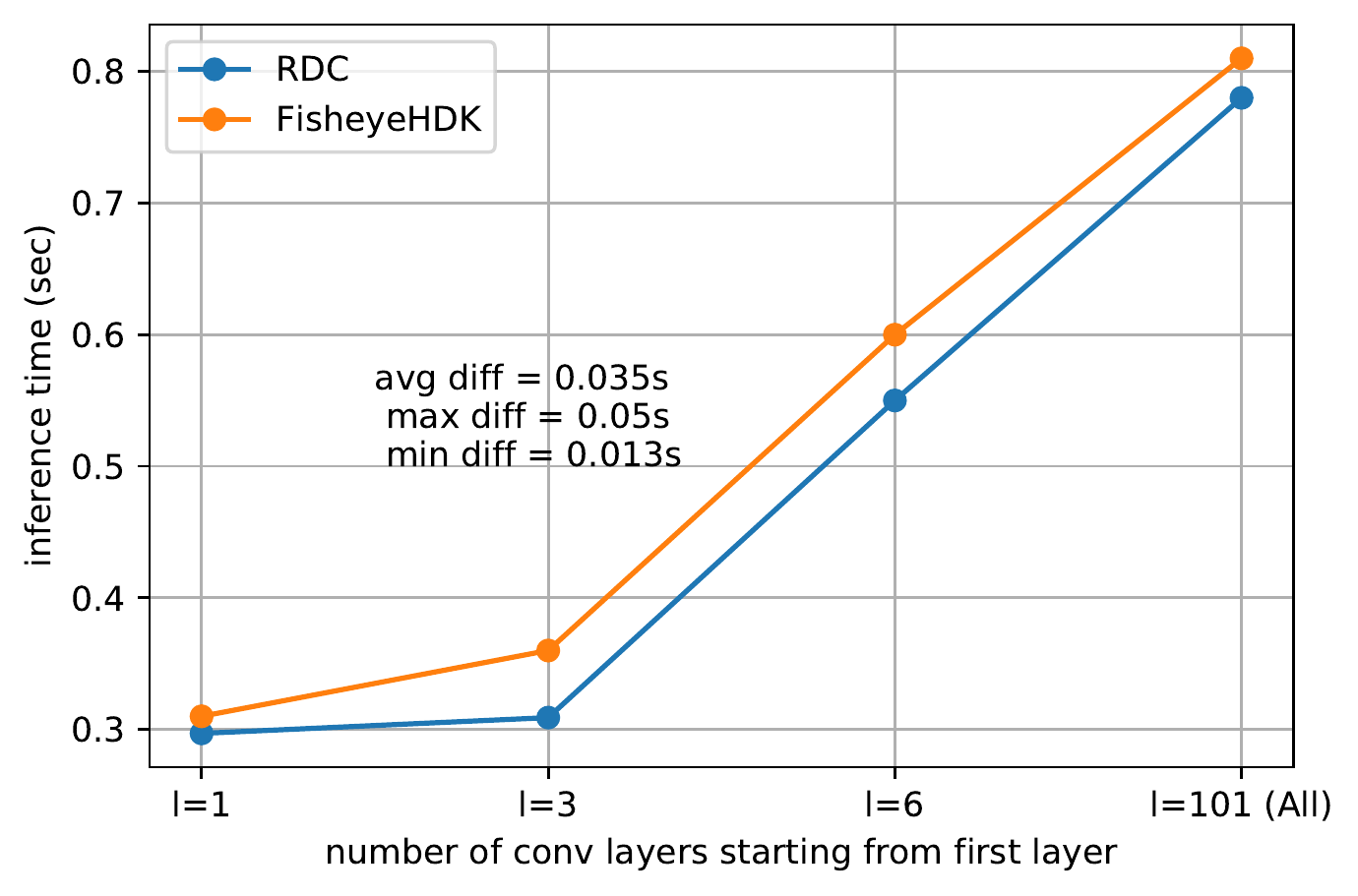} 
\caption{Inference time on single image ($512\times 1024$ pixels)}
\label{fig:time}
\end{figure}

We evaluated the DC approach on synthetically distorted cityscape datasets to compare with both RDC and our method. As can be seen in Table \ref{tbl:f200}, both DC and RDC are comparable. Since DC and RDC are similar as architectures, we use one method and compare the inference time between this method and our method. Fig. \ref{fig:time} shows the inference time for a single image when we implement deformable kernels over first $l$ convolution layers. Table \ref{tbl:suppT} show the interpretation of ResNet-101 layers we used in the main paper.
\begin{table}[t]
\begin{center}
%\resizebox{.95\columnwidth}{!}{
\begin{tabular}{l|l|l}
    $\#$ layers& First layers names & last layers names \\ \hline
    $l=1$ & [conv1] &  [layer4.2.conv3] \\ \hline
    \multirow{ 2}{*}{$l=3$} & [conv1, &  \\
    &layer1.0.conv1,2 & [layer4.2.conv3,2,1]  \\ \hline
     \multirow{3}{*}{$l=6$}& [conv1, &  \\
      &layer1.0.conv1,2,3 & [layer4.2.conv3,2,1] \\
    &layer1.1.conv1,2&  [layer4.1.conv3,2,1]\\
    \hline
    \end{tabular}
    \end{center}

\caption{Interpretation of ResNet-101 layers used in Table 1. of the main paper and in Table \ref{tbl:f200}.}
\label{suppT}
\end{table}
\begin{table}[t]
\begin{center}
\begin{tabular}{lc|p{1.2cm}|p{1.2cm}|p{0.8cm}} %{l |c|c|c}
& $\#$ layers (l) &  DC \cite{8237351}& RDC \cite{8842620}   &  \textbf{ours}  \\
\hline
 \multirow{3}{*}{First layers}&	$l=1$     &57.8			& 57.9	& 58.3 \\ 
 & $l=3$	&	55.6			& 58.0			&58.4 \\
 & $l=6$	&	58.8			& 57.9			&58.5 \\ 
 \hline
 \multirow{3}{*}{Last layers} & $l=6$	&	57.7			& 57.9			&59.3 \\
& $l=3$	&	56.9			&57.7			&59.6 \\
& $l=1$	& 	56.0			&57.9			&58.3 \\ \hline
 & $l=all$	&	55.8			&56.8			&58.0 \\
\hline
\end{tabular}
\end{center}
\caption{Results of mean intersection over union (mIoU (\%)) on distorted cityscapes test set for $f=200$ using ResNet101 backbone module.}
\label{tbl:f200}
\end{table}
\section{Evaluations on BDD100K dataset}
\begin{figure*}
%\begin{center}
 \includegraphics[width=1.0\linewidth]{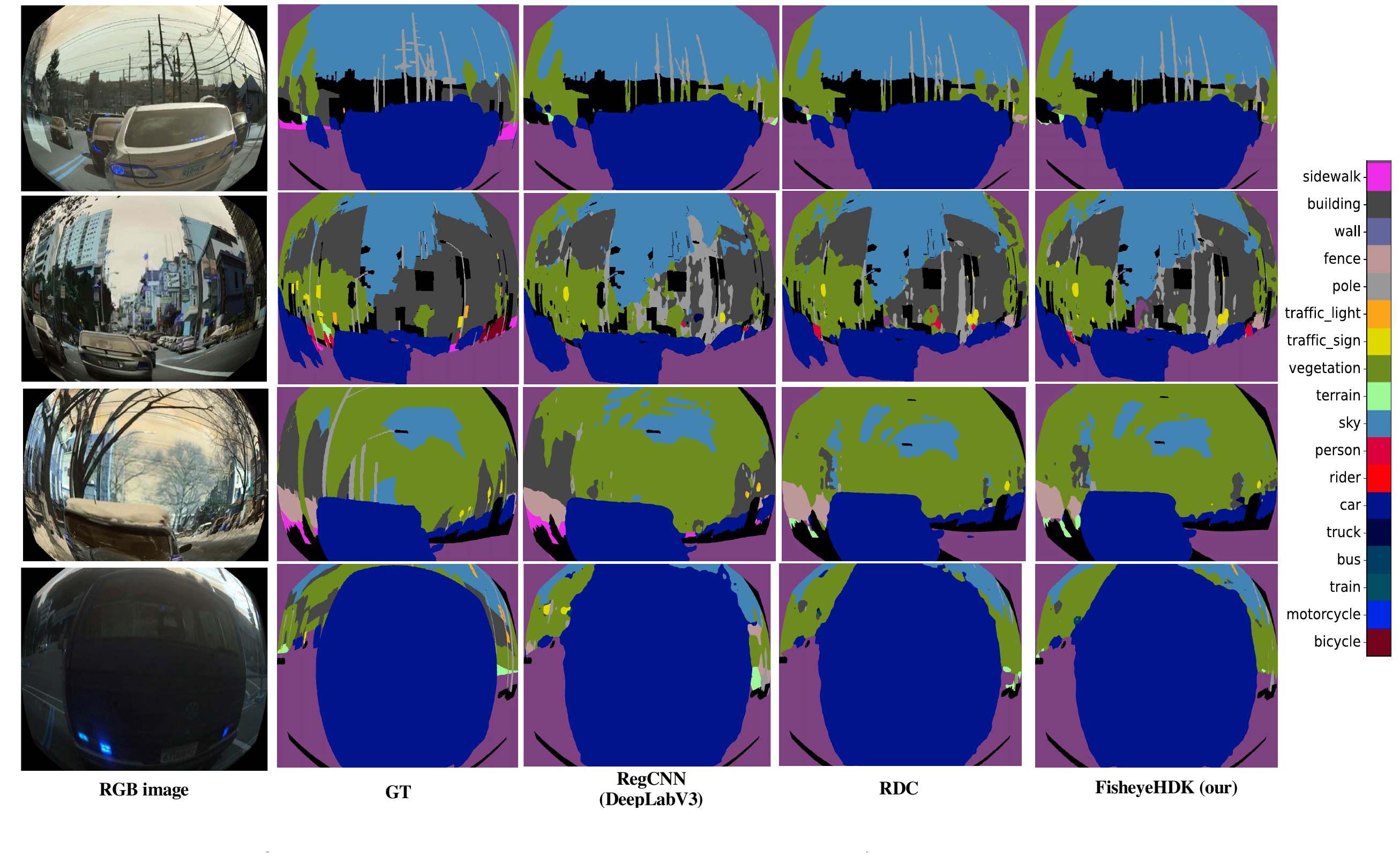}
 %\end{center}
 \caption{Qualitative results of different segmentation methods on BDD100K dataset converted to fisheye with f=75.}
\label{fig:eval_bdd100k}
\end{figure*}

\begin{figure*}
%\begin{center}
 \includegraphics[width=1.0\linewidth]{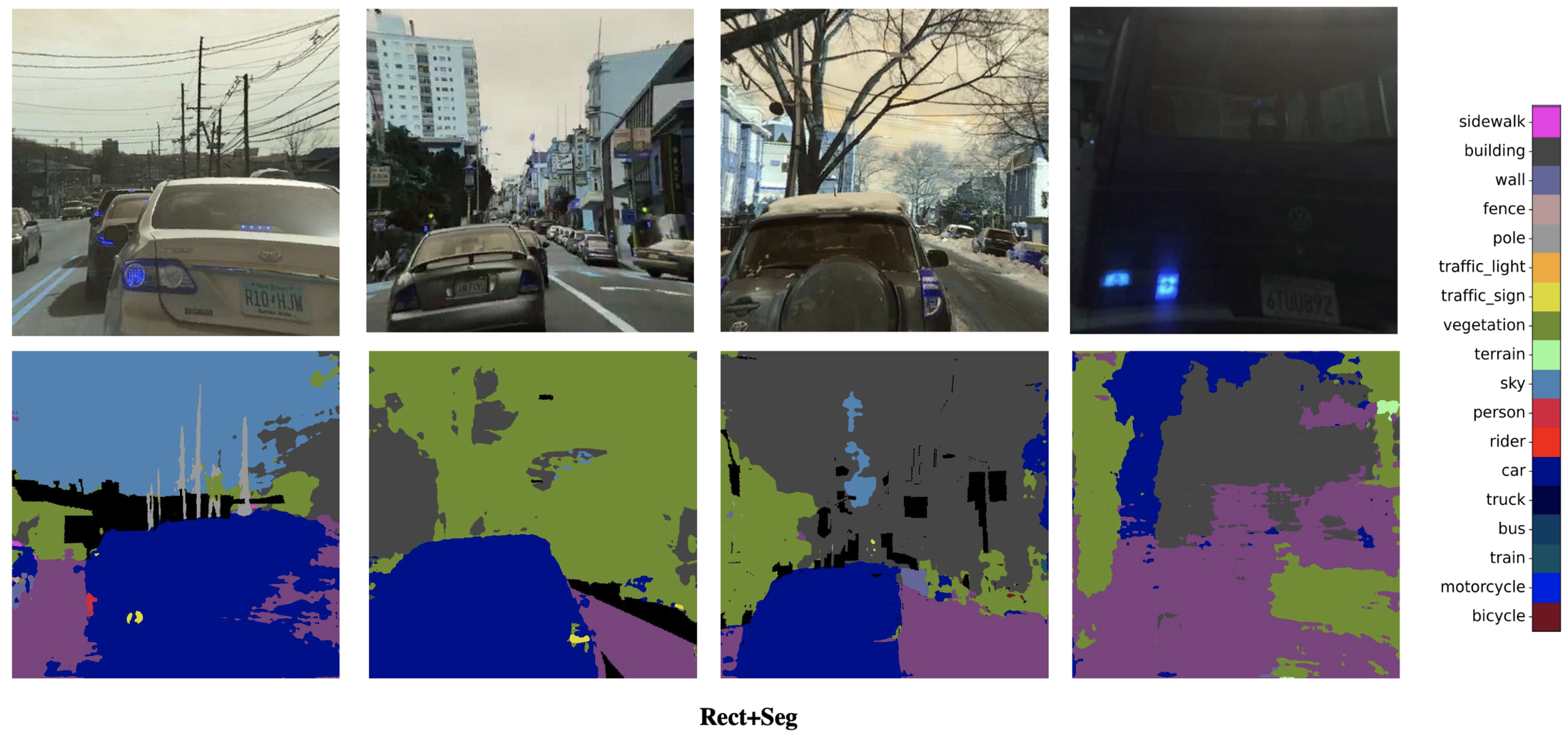}
 %\end{center}
 \caption{Results of rectification then segmentation applied on the same images in figure \ref{fig:eval_bdd100k}.}
\label{fig:rect-seg}
\end{figure*}

We show more qualitative results on BDD100K dataset in Figures \ref{fig:eval_bdd100k} and \ref{fig:rect-seg}. Figure \ref{fig:rect-seg} shows similar images after rectifying fisheye distortion. The segmentation maps are the results of a pre-trained DeepLab model on rectilinear cityscape dataset.
\section{Additional discussions}
We add more discussions and additional suggestions for future improvements. 
\begin{itemize}
\item The HDK network consisted of only one hyperbolic layer. We could increase the number of layers (2 to 3 layers) to explore if deeper network might potentially improve the performance especially in the case of stronger distortions (e.g., f=50). A slightly deeper HDK network will only feed-forward the vertices inputs of the graph across its layers but the adjacency matrix will be used only once after the log map. 
\item Learning deformable kernels in our approach implies using regular graphs in hyperbolic space to predict local displacements inside a kernel at every position in image. We may consider a potential extension of our approach. Instead of only focusing on localized regions inside a deformable kernel, we may use the hyperbolic space to learn a global deformation field as a representation of irregular graph (grid), then sample localized kernels from the deformable grid for convolutions. This extension might enable better generalization to non-linear deformations that need using kernels of multiple shapes and dimensions. Note that all deformable kernel methods typically use a fixed and pre-defined kernel size. 
\item There are crucial benefits of learning features directly from raw large FOV images in real-world applications. Our approach leverages all the information captured from large field of view angles even the details at image's periphery. This is highly important in scene understanding for autonomous driving and in medical imaging such as retinal diagnosis from wide-angle fundus cameras. 
\end{itemize}
%
%
%\newpage

%Bibliography
\bibliographystyle{unsrt}  
\bibliography{suppbib}